\documentclass[lettersize,journal]{IEEEtran}
\IEEEoverridecommandlockouts

\usepackage{cite}
\usepackage{amsmath,amssymb,amsfonts}
\usepackage{hyperref}
\usepackage{cleveref}
\usepackage{graphicx}
\usepackage{tabularx}
\usepackage{textcomp}
\usepackage{xspace}
\usepackage{microtype}
\usepackage{booktabs}
\usepackage{svg}
\usepackage{microtype}
\usepackage{url}
\usepackage{multirow}
\usepackage{arydshln}
\usepackage{pdfpages}

\newcolumntype{C}{>{\centering\arraybackslash}X}
\newcolumntype{L}{>{\raggedright\arraybackslash}X}
\newcolumntype{R}{>{\raggedleft\arraybackslash}X}

\def\BibTeX{{\rm B\kern-.05em{\sc i\kern-.025em b}\kern-.08em
  T\kern-.1667em\lower.7ex\hbox{E}\kern-.125emX}}

\usepackage{pifont} %

\newcommand\etal{\textit{et al.}\xspace}
\newcommand\eg{\textit{e.g.}\xspace}
\newcommand{\ab}{\mathbf{a}}
\newcommand{\ib}{\mathbf{i}}

\newcommand{\ii}[1]{\textcolor{gray}{#1}}
\newcommand{\word}[1]{\textit{#1}}

\usepackage[prependcaption,textsize=tiny]{todonotes}
\definecolor{mycolor}{HTML}{FF6600}
\definecolor{indiagreen}{HTML}{138808}
\definecolor{papaya}{HTML}{EE892F}
\definecolor{mygreen}{HTML}{008000}
\definecolor{mypurple}{HTML}{9966CC}
\definecolor{myblue}{HTML}{5D8AA8}
\definecolor{mypink}{HTML}{EC008C}

\begin{document}

\title{Mapping Written Words to Spoken Words in a Different Language Using Only Visual Grounding}

\author{
    Gabriel Pirlogeanu, Dan Oneata, Horia Cucu, Herman Kamper%
    \thanks{%
        G. Pirlogeanu, D. Oneata, H. Cucu are with the Speech and Dialogue Research Laboratory, \textsc{Politehnica} Bucharest, Romania (email: gabriel.pirlogeanu@upb.ro, dan\_theodor.oneata@upb.ro, horia.cucu@upb.ro).
    }
    \thanks{%
        H. Kamper is with the Department of Electrical and Electronic Engineering, Stellenbosch University, South Africa (email: kamperh@sun.ac.za).
    }
    \thanks{%
    We sincerely thank Rishabh Jain (Trinity College Dublin; rijain@tcd.ie) for his assistance in the manual verification of the Hindi translations.
    }
    \thanks{%
        This work was supported in part by a grants of the Ministry of Research, Innovation and Digitization, CNCS-UEFISCDI, project number PN-IV-P2-2.1-TE-2023-1632, within PNCDI IV.
    }
}

\maketitle

\begin{abstract}
In many low-resource settings, even just eliciting speech for data collection is difficult.
One promising approach has been to ask speakers to describe images.
But how do we build models from such visually grounded speech data?
Given a dataset of images with Hindi spoken captions, %
we %
consider how we can map a %
written English keyword to spoken realisations of that word in
Hindi.
Previous work trained end-to-end multimodal neural models.
Instead, we explore a simpler alignment-based approach built on self-supervised speech representations.
Written English tags are automatically obtained from images using off-the-shelf image captioning systems.
Hindi utterances associated with the same keyword are then aligned (using self-supervised features), and alignment evidence is aggregated to identify recurring speech segments corresponding to the target word.
Experiments evaluating %
keyword spotting and localization show that our alignment-based approach outperforms a previous attention-based neural model.
We also show the benefit of incorporating negative examples during alignment.
Our work demonstrates that cross-lingual word-to-speech mappings can be learned directly from visual grounding without transcriptions or explicit model training.%
\footnote{Code and data are available at \url{https://github.com/gabitza-tech/vgs-cl-vocab}.}

\end{abstract}


\IEEEpeerreviewmaketitle

\section{Introduction}

\begin{figure*}
  \centering
  \includegraphics[width=1\linewidth]{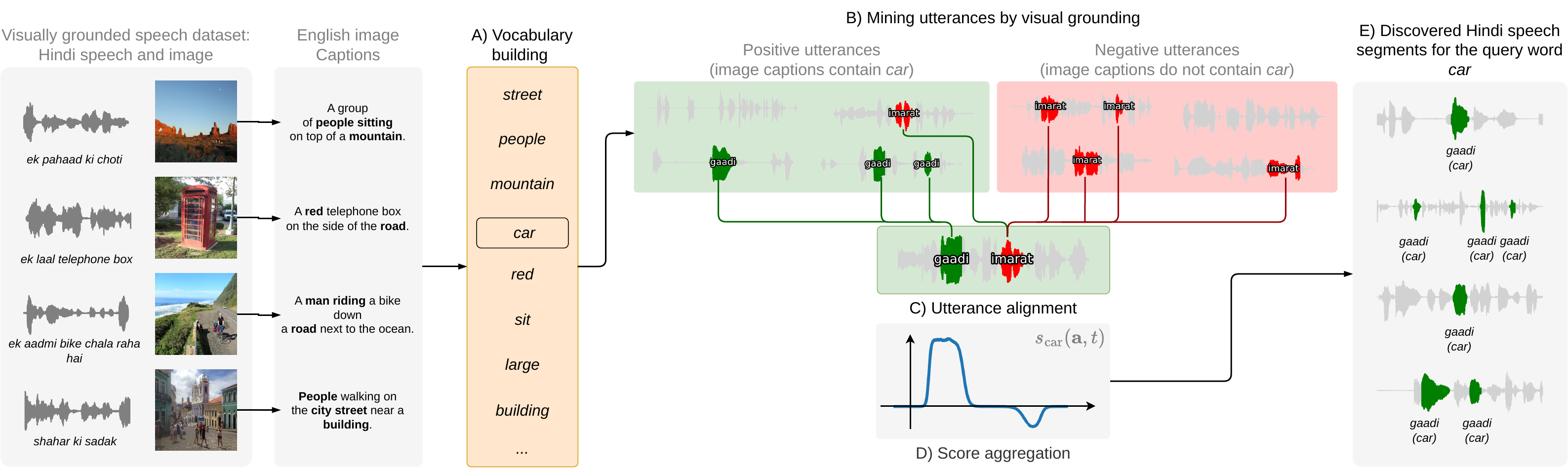}
  \caption{
    Given %
    a visually grounded speech corpus, we 
    learn a mapping between English written words and their spoken Hindi counterparts.
    (A) A vocabulary is constructed from the most frequent words in %
    English captions, obtained automatically from images. (B)
    We partition all utterances based on whether their corresponding image captions contain the target word, \eg, \textit{car}. %
    (C) Then we align a positive utterance with all other utterances and (D) aggregate the alignments:
    positive alignments are added and
    negatives subtracted.
    (E) %
    We retrieve the most aligned Hindi audio segments for a given English keyword. %
    }
  \label{fig:overview}
\end{figure*}

\IEEEPARstart{M}{odern} speech technology has become ubiquitous, yet most of the world's languages remain beyond its reach. %
For many of the world's 7000 
living languages,
even eliciting speech for data collection is challenging because many languages lack a writing systems or are spoken in low-literacy communities.
One promising solution is to collect speech through visual grounding: 
start with a set of images, ask people to describe them, and record their utterances. 
This approach is being adopted by the project Vaani~\cite{vaani,pulikodan2026vaanibenchmarkv10inclusive},
an ongoing effort to preserve the linguistic diversity of India.

In this work we ask whether such a visually grounded speech dataset can be used to document a foreign language.
As a concrete case study, we consider a collection of images paired with Hindi spoken captions and aim to 
(i) discover audio segments corresponding to Hindi words, and
(ii) link the Hindi speech segments to their corresponding English words.
Previous work has explored using visual grounding to map speech to image regions \cite{harwath2018eccv} or even speech to speech across languages~\cite{azuh2019interspeech},
but less emphasis has been placed on mapping speech to written words.
Connecting speech in a foreign language to written words in a high-resource language establishes a clear semantic correspondence and directly supports language documentation.

To obtain textual supervision from images, we propose using off-the-shelf image captioning systems. 
For each image in the dataset, we generate a written description in English using several automatic image captioning systems.
From these captions, we can generate a vocabulary of written words that have a visual correspondence.
To map a keyword to an audio segment, we adapt unsupervised word discovery~\cite{vanniekerk24_interspeech} to incorporate visual grounding.
The idea is to partition the utterances into positive and negative sets based on whether their corresponding image captions contain the target keyword.
A speech alignment method is then run to find the segments that align best with the utterances in the positive set and least with those in the negative set.

Olaleye \etal \cite{olaleye2022jstsp} proposed an approach for the related task
of keyword localization in visually grounded speech:
given an image--speech corpus and a written keyword, locate the keyword in the speech corpus.
Their main idea was to use an image tagger to provide weak supervision for an attention-based audio-to-keyword neural network.
The keywords that can be retrieved are limited and fixed to the codomain of the single image tagger used.
In our case, the vocabulary is dynamically created, based on the corpus at hand.
Moreover, by replying on unsupervised speech segmentation methods for discovering spoken terms,
we are able to more precisely segment the words.

We evaluate two variants for the alignment approach: 
one %
that uses continuous self-supervised features,
while the other relies on discrete features.
Both convincingly outperforms an updated version of the parametric method of Olaleye \etal \cite{olaleye2022jstsp}, with the continuous approach achieving the best overall performance.
Through a series of ablation studies,
we show that our approach is robust to the choice of speech representations, image captioning systems, and hyperparameters.
Since we operate in a cross-lingual setting, a challenge is accounting for differences in how speakers from different cultures perceive the same image.
We provide cross-lingual analyses and contextualize the results against a monolingual setting.

This paper extends our previous work in a conference paper~\cite{pirlogeanu2026connectingspeechwordsimages}.
That study established the core methodology and validated it in a controlled monolingual setting, where both speech and the target keywords were in English.
Here, we move to a more realistic cross-lingual setting by using the Places Hindi dataset \cite{dharwarth2018_interlingua}, in which the speech is in Hindi and the target keywords in English.
In addition, we improve the method by integrating negative information in the alignment process,
and provide more comprehensive analyses, including 
an investigation of the impact of the speech representations and
a detailed word-level error analysis.

\section{Related Work}
\label{sec:related-work}

Our work is related to visually grounded speech models and cross-lingual learning through the visual modality.

\textbf{Connecting Speech to Images.}
A large body of work studies how to connect speech with images, often by learning shared representations between audio and visual data \cite{chrupala2022jair}.
Some methods go further and align parts of speech with parts of an image, such as linking spoken segments to objects or regions in the image \cite{harwath2015asru,harwath2017acl,harwath2018eccv,khorrami2021interspeech}. In these approaches, the model is typically trained to directly align speech features with visual features. In contrast, our method does not rely on learning a fine-grained speech–image alignment model.
Instead, we use images only as a bridge to introduce weak textual supervision, and we rely on %
speech alignment to discover recurring spoken patterns that can later be linked to written words.

\textbf{Connecting Languages via Images.}
Vision provides a common grounding signal across languages, and as such it was used to learn cross-lingual connections. Prior work has shown that visual context can support speech-to-speech retrieval across languages \cite{dharwarth2018_interlingua,havard2019icassp,ohishi2020icassp}, as well as mapping words across languages using images or videos \cite{suris2022cvpr,sigurdsson2020cvpr}.
These studies demonstrate that vision can act as a common reference point between languages.
Particularly relevant to our work is the approach of Azuh \etal~\cite{azuh2019interspeech}, where cross-lingual speech associations are discovered based on two spoken captions of the same image in two different languages.
Differently, we do not assume that we have an English spoken caption, but automatically generate a text description using an image captioning system.

\textbf{Connecting Speech to Words via Images.}
The most related line of work uses images as an intermediate step to connect speech and written words. Early approaches used image tags to define visual concepts and mapped whole speech utterances to sets of words \cite{kamper2018sltu,kamper2019taslp}.
Later work improved this idea by using automatic image captions as richer supervision signals.
Particularly relevant to our work are the findings in~\cite{oneata2024interspeech}, which showed how image captions can guide speech translation, and~\cite{olaleye2022jstsp}, which focused on locating words in speech using image-based supervision. 
We combine these ideas for the first time and show that image captions can also guide localisation of particular keywords in a cross-lingual setting.

\section{Methodology}
\label{sec:methodology}

\subsection{Overview}

We consider a visually grounded speech setting in which images are paired with spoken descriptions in a target foreign language (\eg, Hindi).
Given this data, our goal is to automatically discover recurring acoustic segments and associate them with English written words.

Our approach, illustrated in \Cref{fig:overview}, starts by first defining a vocabulary of English words that can be linked to audio segments (step A).
These are words that have visual grounding and, as such, are likely to be mentioned in the spoken captions.
To obtain the vocabulary, we generate descriptions for the images using pretrained English image captioning models.
The generated descriptions are then lemmatized, and the most frequent concrete words are selected to form the vocabulary (\eg, \textit{street}, \textit{red}, \textit{car}).

Next, given a query word from the vocabulary (\eg, \textit{car}), we want to retrieve corresponding Hindi audio segments.
We propose a new keyword localization method inspired by unsupervised word discovery~\cite{interval_piling,R2018254,dunbar2021zero,vanniekerk24_interspeech},
but which uses the query word to guide the search.
Specifically, we partition the  utterances into positive and negative sets according to whether the query word appears in their associated English image captions (step B; \Cref{sec:filter}).
Each positive utterance, which is likely to contain the query, is aligned with all other utterances (step C; \Cref{sec:align}).
Finally, the alignments are aggregated using an interval piling technique:
alignments with positive utterances are added, whereas alignments between positive and negative utterances are subtracted (step D; \Cref{sec:rank}). 
The highest-scoring segments across the positive utterances are most likely to contain the query word (step E).

\subsection{Mining Utterances Using Visual Information}
\label{sec:filter}

\textbf{Positive Mining.}
Given an audio--image corpus and a query word,
we would like to focus our search on those audio utterances that are likely to contain the query word $w$.
We use the automatically generated image captions as a source of visual supervision:
if the image caption of image $\ib$ contains $w$, then the audio--image pair $(\ab, \ib)$ is included in the positive set:
\begin{equation}
P_w = \left\{ (\ab, \ib)\; \middle|\;  w \in \mathrm{ImageCaptioner}(\ib) \right\}.
\end{equation}
This filtering is shown in \Cref{fig:overview} (step B) in the green left block, where we retrieve all utterances containing the word \word{car} (\word{gaadii}) in the associated image caption.

\textbf{Negative Mining.}
The positive set $P_w$ also covers words that do not match the query word. In one case, you could have words that just very often co-occur with the query, \eg  \textit{road} (\textit{sadak}) often occurs in utterances tagged with \textit{car} (\textit{gaadi}). In other cases, common function words (\eg, \textit{hai}, an auxiliary verb) may dominate.
To avoid such cases, we incorporate contrastive evidence by defining a negative set,
which contains those utterances whose corresponding image captions do not contain the query word:
\begin{equation}
N_w = \left\{ (\ab, \ib)\; \middle|\; w \notin \mathrm{ImageCaptioner}(\ib) \right\}.
\end{equation}
In practice, we use a random subset from $N_w$ with the same size as $P_w$ or a minimum of 50 samples.
Negative sampling is illustrated in the red block of \Cref{fig:overview} (step B), 
where we retrieve utterances that do not contain the word \word{car} (\word{gaadii}) in their captions.

\textbf{Semantically Negative Mining.}
To directly target co-occurrences, we also define a set of negative utterances that specifically contain words that co-occur with the query $w$:
\begin{align}
N'_w = \{ (\ab, \ib)\; | \;
&\mathrm{IsCooccurence}(c, w), \notag\\
&c \leftarrow \mathrm{ImageCaptioner}(\ib), \notag\\
&(\ab, \ib) \leftarrow N_w \}.
\end{align}
A word $c$ is considered to co-occur with the vocabulary word $w$ if it is one of the two most frequent co-occurring words with $w$.
As with the negative set, we use a random subset from $N'_w$ with the same size as $P_w$ or at a minimum of 50 samples.

\subsection{Aligning Utterances}
\label{sec:align}

\begin{figure}[t]
  \centering
   \includegraphics[width=0.8\linewidth]{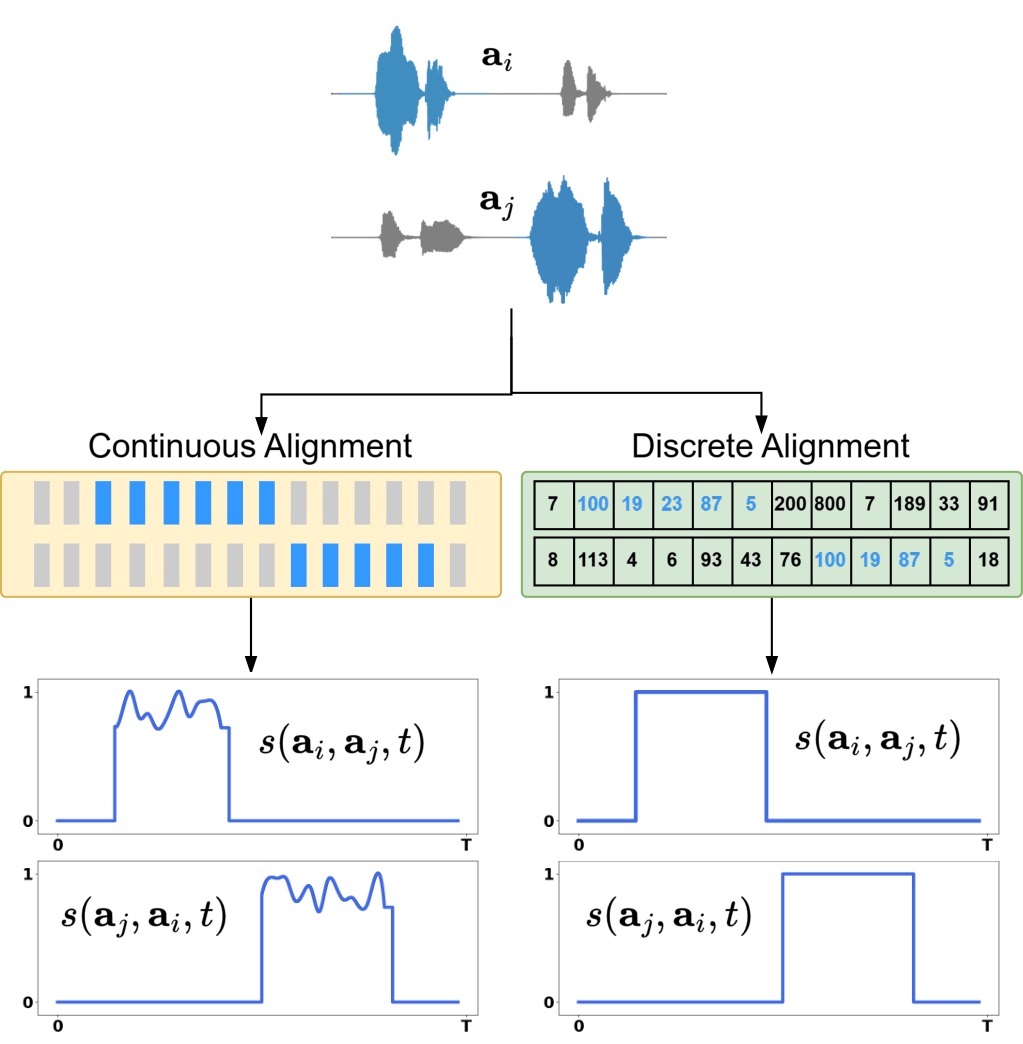}
  \caption{
    Continuous and discrete alignment methods are applied to a pair of input utterances.
    The discrete method takes cluster codes as input and produces a binary alignment signal. The continuous approach yields %
    alignment
    scores in the range \([0, 1]\).
  }
  \label{fig:comparison_methods}
\end{figure}

Next, we align the utterances with each other.
The goal of this step is to identify recurring audio subsequences that consistently appear in the corpus.
The query word will then be found among these subsequences.
We extract the common subsequences using an unsupervised word discovery approach. This is illustrated in~\Cref{fig:overview} (step C), where we align each positive utterance with all other positive utterances, as well as with all negative utterances.

Formally, for
any two audio utterances $\ab_i$ and $\ab_j$, we define a scoring function $s(\ab_i, \ab_j, t)$ for %
how likely it is that segment $t$ from %
utterance $\ab_i$ appears anywhere in %
utterance $\ab_j$. 
We explore two variants:
one based on continuous self-supervised representations,
and the other on discrete ones.
Both variants rely internally on HuBERT representations \cite{hubert2021} and are illustrated in \Cref{fig:comparison_methods}.

\textbf{Discrete Features Alignment (DFA).}
As in \cite{vanniekerk24_interspeech},
HuBERT features are first encoded to discrete units using k-means clustering, and then each pair of unit sequences are
aligned using the Smith-Waterman dynamic programming algorithm~\cite{smith-waterman}.
From the resulting alignment, we construct a binary scoring function $s$:
1 if the segment in $\ab_i$ is matched, and 0 otherwise.
A similarity threshold $\tau$ sets a
minimum %
score required for a pair of sub-sequences to be considered a match, thereby controlling the trade-off between the quantity and quality of the discovered patterns.%

\textbf{Continuous Features Alignment (CFA).}
Following \cite{azuh2019interspeech}, we estimate the alignment of two utterances by computing the similarity directly in the feature space.
We use cosine similarity between features: we  %
define the score between a pair of utterances %
as the maximum similarity between the features $\phi_{it}$, extracted from the $\ab_i$ at time $t$, and all the other features $\phi_j$ from the second utterance: %
\(s(\ab_i, \ab_j, t) = \max_{t'} \langle \phi_{it}, \phi_{jt'} \rangle\).
To obtain a less noisy and sparse signal,
we apply a smoothing filter (using a standard Gaussian) and set the values below \( \gamma \cdot \max (s) \) to 0, where $\gamma$ is the local threshold hyperparameter.
While continuous features have more capacity than discrete ones, framewise comparisons are much slower.

\subsection{Ranking Segments by Aggregation}
\label{sec:rank}

Given the alignments between a positive audio file and the rest of the utterances,
we can locate the query word.
The idea is that the query is likely to appear in the subsequences shared with the other positive utterances,
and is likely to be absent from those shared with the negative utterances.
This is illustrated in \Cref{fig:overview} (step D), where we show the score for a given audio sample, $\ab$, while searching for the word \word{car}. As noted, we expect the queried word to be well-aligned, whereas co-occurring terms should be penalized through negative mining.

Concretely,
given a positive sample $\ab$,
we define the utterance-wise score by
summing its alignment scores with all other positive samples $\ab'$ and subtracting its alignment scores with all negative samples $\bar\ab$:

\begin{equation}
s(\ab, t) =
\sum_{\substack{\ab\ \in P_w \\ \ab' \neq \ab}}
s(\ab, \ab', t)
\;-\;
\sum_{\bar\ab \in N_w}
s(\ab, \bar\ab, t).
\end{equation}

This formulation extends the ``interval piling''~\cite{interval_piling} approach with a contrastive component: alignments across positive pairs increase the score, while alignments with negative samples are penalized.
This suppresses frequent but non-informative segments such as co-occurrences and function words.
For the negative set we can also use the semantic negative set $N'_w$;
we also consider a no-negatives (only-positives) variant as a baseline. The benefits of negative samples are explicitly investigated in our experiments.

To eliminate noisy alignments, we discard values below a global threshold $\theta$, defined as a fraction of the maximum aggregated score within the utterance.
The speech is then segmented in contiguous regions delimited by silence.
Each resulting segment is scored using the average of its frame-level values.
The high-scoring segments are selected as candidate instances of the query word spoken in the foreign language.

\section{Experimental Setup}

\subsection{Data and Groundtruth}
\label{sec:vgs-data}

We use the MIT Place Audio Captions Hindi dataset \cite{dharwarth2018_interlingua}, a speech--image dataset %
of 100k images paired with spoken Hindi descriptions.
The images come from the Places dataset \cite{NIPS2014_19ea3982}, and contain scenes such as forests, kindergarten classrooms, or car interiors.
The spoken captions were collected via Amazon Mechanical Turk and are spontaneously spoken rather than read.
For our experiments, we sample 20k speech--image pairs and use 10k %
for development purposes and the other 10k for the final evaluation.
The selected samples are at most seven seconds long;
this improves alignment, while still preserving enough content.
For comparative monolingual experiments, we also use the English variant of the dataset, Places Audio Captions \cite{harwath2016nips,harwath2017acl,harwath2018eccv}, which is similarly constructed, but with English spoken captions;
we use
the spoken English descriptions for the same 20k images as for the Hindi set.

To evaluate our approach, we need to know where an English word appears in a Hindi spoken caption.
This requires Hindi transcriptions, word-level forced alignments, and English--Hindi translations.
The original dataset
does not provide these,
 so we construct the groundtruth ourselves.
First, we obtain automatic Hindi transcriptions using the Conformer Large CTC model \cite{conformer2021} from NVIDIA NeMo~\cite{kuchaiev2019nemo};
this model achieves a word error rate of 9.4\% on the MUCS 2021 dataset~\cite{diwan21_interspeech}.
The transcriptions are then aligned using a CTC-based forced aligner based on the Massively Multilingual Speech~\cite{scalingspeech2024} model.
Finally, the transcripts are romanized using the \texttt{uroman} library~\cite{hermjakob-etal-2018-box} and lemmatized using Stanza~\cite{qi-etal-2020-stanza}.
This enables the use of a simple bilingual dictionary that maps English keywords to romanized Hindi lemmas.
The dictionary is obtained with ChatGPT (as well as manually checked)
and is one-to-many, including multiple valid translations
(\word{car} $\rightarrow$ \{\word{gaadi}, \word{kaar}\})
and near synonyms
(\textit{road} $\rightarrow$ \{\word{sadak} (\word{road}), \word{gali} (\word{alley})\}).

\subsection{Implementation Details}
\label{subsec:implementation}

We generate image decriptions using three image captioning systems: Tag2Text~\cite{huang2023tag2text}, BLIP-2~\cite{10.5555/3618408.3619222}, and GIT~\cite{wang2022gitgenerativeimagetotexttransformer}.
For each image, captions are produced via beam search and then processed to remove stopwords and lemmatized with the \texttt{en\_core\_web\_sm} SpaCy model~\cite{Honnibal_spaCy_Industrial-strength_Natural_2020}.
The final set of words associated with each image is the intersection of the words produced by all three captioners for that image.
These words are used to ensemble the vocabulary and for the utterance selection step (\Cref{sec:filter}).
The vocabulary contains the most frequent 100 words (\eg, \word{chair}, \word{red}, \word{sit}),
except for visually ungrounded terms (\eg \textit{background},  \textit{picture}, \textit{view}), which were manually removed.
The vocabulary is constructed independently on both development and test splits.

For both alignment methods, we extract features from the seventh layer of the English HuBERT Base model, the optimal layer for phone discrimination~\cite{hubert2021}.
To prevent the alignment of silence or background noise across utterances, we apply voice activity detection using Pyannote3~\cite{Plaquet23}.
Additionaly, {the pipeline's postprocessing} hyperparameters prevent short segments:
we remove {segments resulted from pair-wise alignment that are shorter} than a \textit{min. duration local} value;
aggregated segments are discarded if shorter than a \textit{min. duration global} value;
{we add padding at the start of retrieved segments with a \textit{pad\_on} value;
and also pad them at the end with a \textit{pad\_off} value.}
All hyperparameters are tuned in the monolingual setting, on the English development set, and the exact values are given in \Cref{tab:hyperparams}.
We observe that the two sets of hyperparaters are similar across the two alignment variants. Moreover, they are also similar to what we obtain if we were to tune on the Hindi development set (results not shown here).
This confirms the robustness of the approach.

\begin{table}[t]
  \centering
  \caption{%
  Hyperparameters selected on the monolingual
English development set for the two alignment methods.
  }
  \begin{tabularx}{\linewidth}{Xrrrrrrr}
  \toprule
   & & & & \multicolumn{2}{c}{min. duration} \\
   \cmidrule(lr){5-6}
   Method & $\tau$ & $\gamma$ & $\theta$ & local & global & pad\_on & pad\_off \\
  \midrule
  CFA & \color{gray}{N/A} & 0.7 & 0.6 & 0.2 & 0.2 & 0.0 & 0.1 \\
  DFA & 3 & \color{gray}{N/A} & 0.7 & 0.2 & 0.2 & 0.0 & 0.1 \\
  \bottomrule
  \end{tabularx}
  \label{tab:hyperparams}
\end{table}

\subsection{Evaluation Protocol}
\label{sec:eval}

We evaluate our models in terms of two retrieval metrics: keyword localization and keyword spotting.
For keyword localization, a retrieved segment for a given keyword is considered correct if it overlaps sufficiently with a Hindi translation of that keyword;
overlap is computed in terms of the intersection over union (IoU) of the two segments and it has to exceed a predefined threshold (we use 0.5 throughout) %
For keyword spotting, a retrieved segment is considered correct if the Hindi translation of the query keyword appears anywhere in the utterance;
this metric is a strict upper-bound on the localization performance.
Since the mapping is one-to-many,
a retrieval is counted correct if it matches any of the valid translations.
For both localization and spotting, we take the top 10 retrieved audio segments for each word in the vocabulary and report the number of correct predictions (P@10). 
The scores are averaged over the vocabulary words. %

\subsection{Alternative Approach and Toplines}
\label{sec:baseline}

\textbf{Attention CNN.}
For comparison, we consider our own updated version of the %
neural-based model of
Olaleye \etal \cite{Olaleye2021AttentionBasedKL, olaleye2022jstsp}, the only other work to look at a related task.
The model has two inputs, an audio utterance and a word from the vocabulary, and predicts whether that word appears anywhere in the utterance (based on the associated image captions).
We use HuBERT features from the seventh %
transformer layer as input, to ensure a fair comparison with our method.
The model
consists of
a convolutional neural network, an attention layer (to pool the audio embeddings over time, based on the input word),
and a two-layer perceptron (to project the pooled embedding to a binary prediction).
To localize 
a word, we find the peak of the attention weights and return a fixed-sized segment around the peak (from 0.1s before to 0.3s after the peak).
This is necessary since the attention plots are very peaky, resulting in poor performance for this approach on our benchmark (\cite{Olaleye2021AttentionBasedKL} only evaluated whether the peak occurred within the true word, not whether a segment can be extracted).
The displacement hyperparameters were tuned on the English development set.

\textbf{Transcript Topline.}
Image captions only approximate the words in an utterance.
To understand the system's potential under ideal conditions, we evaluate its performance using the utterance's actual transcript.
Specifically, we carry out the selection step (\Cref{sec:filter}) by checking whether any of the translations of the English word appear anywhere in the Hindi transcript.

\textbf{Monolingual Topline.}
We also consider a monolingual setting in which both the query word and the target spoken language are English.
This setting is still challenging because, although the spoken captions are in English, they are not perfectly aligned with the generated image captions.
Nevertheless, it serves as an upper bound for our main cross-lingual setting, as it avoids cultural biases in the image descriptions and uses speech representations that are optimized for English.

 \section{Results}
\label{sec:experimental-results}

\subsection{Main Results}
\label{sec:hindi-results}

\begin{table}[t]
  \centering
  \caption{%
  P@10 (\%) for cross-lingual keyword spotting and localization using continuous (CFA) or discrete (DFA) feature alignment. %
  Toplines use transcripts; visually grounded systems use the intersection of three image taggers.
  }
  \begin{tabularx}{\linewidth}{Xlrr}
  \toprule
  Method & Mining & Spotting & Localization \\
  \midrule
  \multicolumn{4}{l}{\it Toplines: Use transcripts for supervision} \\
  \addlinespace
  \multirow[c]{3}{*}{DFA}
                  & pos                     & 100 & 40.8 \\
                  & pos \textit{vs} neg     & 100 & 69.4 \\
                  & pos \textit{vs} sem neg & 100 & 71.8 \\
  \addlinespace
   \cdashline{2-4}
  \addlinespace
  \multirow[c]{3}{*}{CFA}
                  & pos                     & 100 & 75.7 \\
                  & pos \textit{vs} neg     & 100 & \textbf{89.8} \\
                  & pos \textit{vs} sem neg & 100 & 89.2 \\
  \midrule
  \multicolumn{4}{l}{\it Visually grounded systems: Use images for supervision} \\
  \addlinespace
  Attention CNN\cite{olaleye2022jstsp} & \color{gray}N/A & 18.8 & 10.4 \\
  \addlinespace
   \cdashline{2-4}
  \addlinespace
  \multirow[c]{3}{*}{DFA}
                  & pos                     & 49.7 & 16.8 \\
                  & pos \textit{vs} neg     & 56.8 & 34.2 \\
                  & pos \textit{vs} sem neg & 55.7 & 34.6 \\
    \addlinespace
   \cdashline{2-4}
  \addlinespace
    \multirow[c]{3}{*}{CFA}
                  & pos                     & 47.7 & 23.6 \\
                  & pos \textit{vs} neg     & \textbf{63.0} & \textbf{49.9} \\
                  & pos \textit{vs} sem neg & 61.5 & 47.5 \\
  \bottomrule
  \end{tabularx}
  \label{tab:main-results-hindi}
\end{table}

\begin{figure*}
  \centering
  \setlength{\tabcolsep}{0pt}
  \begin{tabular}{@{}c@{}c@{}c@{}c@{}}
  \includegraphics[width=0.25\textwidth]{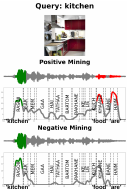} &
  \includegraphics[width=0.25\textwidth]{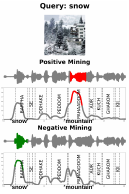} &
  \includegraphics[width=0.25\textwidth]{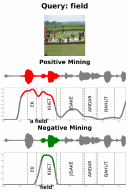} &
  \includegraphics[width=0.25\textwidth]{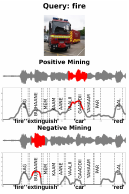}
  \end{tabular}
  \caption{%
    Examples of the top retrieved segments for four query words with the baseline filtering mechanism (top) and with negative mining (bottom).
    CFA is used for audio alignment.
    For better visualization, we overlap the forced alignments over the aggregated signal plot.
    Images correspond to the utterances below. 
  }
  \label{fig:four-images}
\end{figure*}

\Cref{tab:main-results-hindi} presents our main results.
We consider two settings:
a topline setting in which transcripts are used for supervision, and
the actual visually grounded models in which models are supervised directly with images.
In each setting, we compare the two proposed alignment variants
(continuous feature alignment, CFA; discrete feature alignment, DFA; Section ~\ref{sec:align}),
and the three ways of mining variants
(positives only; positives and negatives; positives and semantic negatives; Section ~\ref{sec:filter}).
In the realistic visually grounded setting, we also compare to the attention-based baseline (Attention CNN; Section~\ref{sec:baseline}).
We now summarize our main findings.

\textbf{Continuous Features Perform Best.}
Across all settings, alignment based on continuous features (CFA) consistently achieves the strongest results.
For example, 
the best visually grounded localization result is 49.9\% P@10, 
obtained by CFA with positive and negative selection;
DFA achieves only 34.2\% in the same setting.
This gap likely arises because continuous features retain more information 
since information is not discarded through a discretizing clustering step.
This advantage, however, comes at an efficiency cost:
aligning 250 audio clips takes 2m15s with CFA, compared to only 24s with DFA.
Importantly, both CFA and DFA outperform the Attention CNN method \cite{olaleye2022jstsp},
thereby improving on the current state-of-the art.

\textbf{Negative Mining Consistently Helps.}
Incorporating contrastive information through the negative samples helps across all settings,
for both CFA and DFA and for both the toplines and visually grounded systems.
For example,
in the visually grounded localization task,
DFA improves from 16.8\% to 34.6\%, and CFA from 23.6\% to 49.9\%.
Restricting negatives to semantic negatives yields similar, but slightly weaker performance than using unrestricted negatives.
Interestingly, negatives have a greater impact on localization than on keyword spotting:
23.6\% to 49.9\% (a 111\% relative improvement) versus
47.7\% to 63.0\% (a 32\% relative improvement).
This is because negatives suppress co-occurrening words, which primarily affect localization, not keyword spotting.

\Cref{fig:four-images} illustrates this qualitatively.
For the query \word{kitchen} (\word{rasoi}), negative mining suppresses the alignment with the auxiliary word \word{are} (\word{hain}) and the semantically related word \word{food} (\word{khaana}).
In the second example, negative mining helps correctly retrieve the word \word{snow} (\word{barf}) by reducing the alignment with the frequently co-occurring word \word{mountain} (\word{pahaad}).
The third example shows an improvement in localization, removing the alignment with the neighbouring numeral \word{one} (\word{ek}).
However, co-occurrence errors can still persist. In the fourth example, the query \word{fire} (\word{aag}) is aligned with \word{car} (\word{gaadi}) when using positive mining, and with the verb \word{extinguish} (\word{bujhaane}) when using negative mining. Since most images depict \word{fire trucks} or \word{fire stations} together, the concept of \word{fire} is associated with many co-occurring words such as \word{car}, \word{red}, \word{extinguish}, and \word{parking}.

\textbf{Weak Supervision is the Primary Source of Performance Loss.}
For the transcript topline, we are guaranteed that a word is present in the utterance.
Under this idealized scenario, results are very strong:
spotting is perfect, localization performance is approaching 90\% with CFA (top section of Table~\ref{tab:main-results-hindi}).
This suggests that the proposed methodology (alignment and utterance mining) performs well as long as the data is not noisy.
The invariable mismatch between what is spoken and what is seen is what degrades performance.
This mismatch is not unique to machine-generated captions:
it also affects human-generated ones \cite{wang2016lrec},
since two people may describe the same image using different words.
In our case, the issue is further compounded by cultural and linguistic gaps.
We analyse this aspect further next. %

\subsection{The Cross-Lingual Gap}
\label{sec:cross-lingual-gap}

\begin{table}[t]
  \centering
  \caption{%
  P@10 (\%) %
  for keyword spotting and localization in the monolingual (English--English) and cross-lingual (Hindi--English) settings. We use continuous alignment (CFA) with negative mining.
  }
  \setlength{\tabcolsep}{5pt}
  \begin{tabularx}{\linewidth}{@{}rXllrr}
  \toprule
  & & \multicolumn{2}{c}{Language} \\
  \cmidrule(lr){3-4}
  & Setting & Utterances & Keywords & Spotting & Localization \\
  \midrule
  & \multicolumn{5}{l}{\it Toplines: Use transcripts for supervision} \\
  \ii{1} & Mono-lingual  & English & English & 100 & 94.3 \\
  \ii{2} & Cross-lingual & Hindi   & English & 100 & 89.8 \\
  \midrule
  & \multicolumn{5}{l}{\it Visually grounded systems: Use images for supervision} \\
  \ii{3} & Mono-lingual  & English & English & 86.3 & 75.4 \\
  \ii{4} & Cross-lingual & Hindi   & English & 63.0 & 49.9 \\
  \bottomrule
  \end{tabularx}
  \label{tab:comparison}
\end{table}

As we saw, the inherent variation in how an image is described makes the studied task challenging.
This gaps widens further in a cross-lingual setting.
Since captioning models are trained on English-centric data from Western culture, while the spoken captions come from speakers based in India,
concepts of interest to a Westerner, such as \textit{golf}, might not be referred to at all by a native of India \cite{hershcovich2022challenges}.
Here
we analyse the challenges that arise from the cross-lingual and cultural gap.

\textbf{Cross-linguality Partially Explains the Performance Gap.}
To quantify the performance drop due to working with a foreign language, we compare our default cross-lingual setting to a monolingual setting, in which both the utterances and the keywords are in English.
The monolingual English--English setting places %
an upperbound on performance, since English is also the language on which the image captions and speech representations are trained.
\Cref{tab:comparison} reports results in both idealized (transcript-based) and visually grounded (image-based) settings for the two languages.
We observe that, in the monolingual setting, the visually grounded model partially closes the gap to the cross-lingual topline:
localization %
improves from 49.9\% (row 4) to 75.4\% (row 3), with the topline being at 89.8\% (row 2).
A gap still remains, due to the imperfect supervision provided by image captions.
We also observe that the cross-lingual topline is %
close to the mono-lingual one:
89.8\% (row 2) and 94.3\% (row 1).
Since the idealized setting removes the effect of image captioning,
we conclude that the language of the speech representations---the only remaining difference---is not a crucial factor.
But we do formally
analyse the impact of speech representations in \Cref{subsec:speech_repr_analysis}.

\begin{table}[t]
  \centering
  \caption{%
    Alignment between captions generated either automatically or manually by English or Hindi annotators.
    Alignment is measured in terms of precision %
    and recall %
    and averaged over all 10k samples in development set and 100 words in the vocabulary.
  }
  \begin{tabularx}{\linewidth}{@{}lXXrr}
  \toprule
  & Prediction & Groundtruth & Prec.\ (\%) & Rec.\ (\%) \\
  \midrule
  \ii{1} & Automatic (English) & English annotators & 50 & 47 \\
  \ii{2} & Automatic (English) & Hindi annotators   & 26 & 40 \\
  \ii{3} & English annotators  & Hindi annotators   & 22 & 37 \\
  \bottomrule
  \end{tabularx}
  \label{tab:alignment-results}
\end{table}

\textbf{English and Hindi Speakers Describe Images Differently.}
To measure the cross-lingual gap directly, we compare Hindi and English captions provided for the same set of images by human annotators.
An English keyword is said to occur in the Hindi caption if one of its translations appears in the lemmatized Hindi caption.
To contextualize the results, we also compare automatically generated captions with captions from English or Hindi annotators.
\Cref{tab:alignment-results} shows precision and recall on the 10k images from the development set, averaged over the 100 keywords in the vocabulary.
We observe moderate alignment in the English--English setting between automatic captions and manual ones.
Alignment drops in the cross-lingual setting, with the larger drop being in precision (words used in English captions do not occur in the Hindi ones).
Interestingly, Hindi annotations align better with automatic captions (row 2) than other English annotators (row 3),
confirming that the issue does not lie with the captioning system, but is more fundamental in the way that humans describe the images.

We investigate particular cases and observe that systematic differences appear for some categories.
One such case is sports:
English speakers naturally name the specific sport being depicted, \textit{baseball}, \textit{golf}, or \textit{bowling}.
Hindi speakers tend to
conflate them under a single category---\eg, \textit{game} (\word{khel})---or do not name the sport at all, instead referring to other parts of the image, \eg \textit{field} (\word{maidaan}) or %
\textit{players} (\word{khiladi}).
A similar pattern emerges for scene-level concepts:
instances that might be labelled a \word{castle} (\word{kilaa}, \word{mahal}) in English are more commonly described in Hindi using broader terms such as \word{old/ancient building} (\word{purani imaarat}) or \word{structure} (\word{sanrachna}).
Instead of specifically naming a scene \word{restaurant},
Hindi speakers often refer to its constituent elements, \eg, \word{table} (\word{mez}) or \word{chair} (\word{kursi}). %

\section{Further Analyses}
\label{sec:ablations}

To further understand our system,
we investigate how the choice of
image captioning systems (\Cref{subsec:captions-perf}) and
speech representations (\Cref{subsec:speech_repr_analysis})
impact the final performance.
At the end, we take a more granular look and inspect how the keyword-level performance is impacted by factors such as number of samples or co-occurrence percentage (\Cref{subsec:errors-word}).

\subsection{Impact of Image Captioning Systems}
\label{subsec:captions-perf}

\begin{figure}[t]
  \centering
  \includegraphics[width=0.43\textwidth]{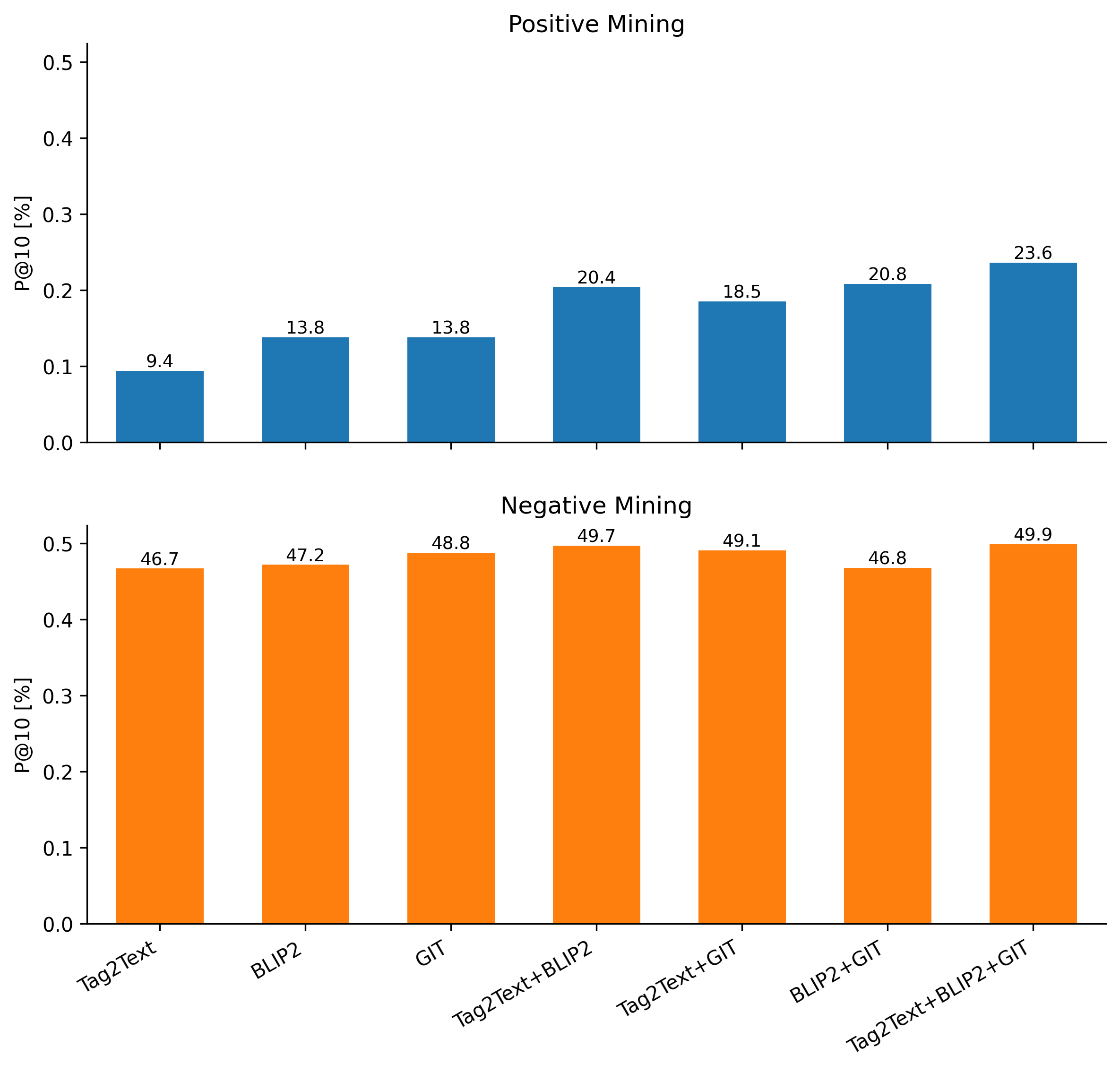}
  
  \caption{Localization performance P@10 on the Hindi test set using CFA as the captioning system for image-based supervision is changed.
  The top plot uses positive mining as the filtering mechanism, while the bottom plot uses negative mining.}
  \label{fig:results_plots}
\end{figure}

Image captions are used in the mining stage to select positive and negative utterances.
We now examine how the choice of captioning systems affect downstream performance.
For the main experiments we used the intersection of three captioning systems.
Here, we report performance when captions are taken from a single system or from the intersection of any two pairs of systems.

\Cref{fig:results_plots} shows results for positive and negative mining strategies across all combinations of captioning systems.
For the proposed negative mining variant, performance is largely insensitive to the underlying captioning system:
all captioning combinations give very similar---and strong---results (between 46.7\% and 49.9\%).
In contrast, the baseline positive mining strategy is strongly affected by the choice of captioning system.
Combining multiple systems tends to have a positive impact, increasing localization performance from 13.8\% P@10 with the best single captioner, BLIP2, to 23.6\% with the intersection of three.
This is likely because the intersection of captions
produces more precise annotations:
captioning precision increases from 33.37\% (best single captioner) to 41.5\% (intersection of three captioners).
As we show later (\Cref{subsec:errors-word}),
precision at the word-level also correlates positively with performance.

\begin{table}[t]
  \centering
  \caption{
  P@10 (\%) for localization on the Hindi test set when varying the self-supervised speech representations:
  model architecture, feature extraction layer, and pretraining language. Results use the CFA method with negative mining.
  }
  \begin{tabularx}{\linewidth}{
  Xrlrr
                }
 
  \toprule
  Architecture & Layer & Language & Hours & P@10 \\
  \midrule
  \multicolumn{5}{l}{\textit{Default configuration}} \\
  HuBERT Base~\cite{hubert2021}      & 7 & English & 960 & 49.9 \\
  \midrule
  \multicolumn{5}{l}{\textit{Other configurations}} \\
  HuBERT Large~\cite{hubert2021}      & 24 & English & 60k & 38.5\\
  HuBERT Large~\cite{hubert2021}      & 24 & Mandarin & 10k & 36.7 \\
  mHuBERT~\cite{boito2024mhubert} &  8 & Multilingual & 90k & \textbf{50.7} \\
  WavLM Base~\cite{Chen2021WavLMLS}  &  8 & English & 960 & 45.4 \\
  WavLM Large~\cite{Chen2021WavLMLS}  & 24 & English & 94k & 39.4 \\
  wav2vec 2.0 Base~\cite{wav2vec2}          &  6 & English & 960 & 36.7 \\
  wav2vec 2.0 Base~\cite{wav2vec2}          &  6 & Hindi & 10k & 44.4 \\
  \bottomrule
  \end{tabularx}
  \label{tab:speech_repr}
\end{table}

\begin{figure*}
  \centering
  \includegraphics[width=0.9\linewidth]{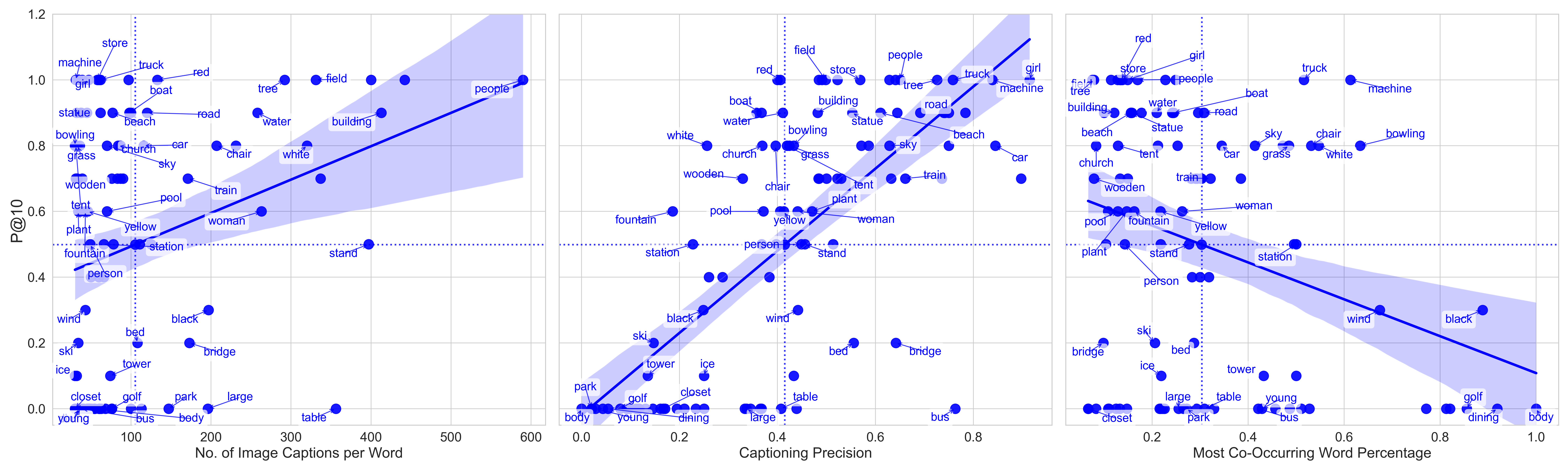}
  \caption{
    Word-level performance (P@10 at IoU = 0.5) as a function of three factors:
    (left) number of image captions for each word;
    (middle) captioning precision for each word;
    (right) percentage of times the best co-occurrence co-occurs with each word.
    Results are obtained using CFA with negative mining on the Hindi test set.
    Each plot shows a linear regression fit and the 95\% confidence interval computed by bootstrapping.
    Dotted lines %
    represent mean values.
    }
  \label{fig:plots}
\end{figure*}

\subsection{Impact of Speech Representations}
\label{subsec:speech_repr_analysis}

The speech representations are critical for the alignment step (\Cref{sec:align}).
Here, we investigate multiple options for speech representations by
considering models of different scales, trained on different data and different training languages.

\Cref{tab:speech_repr} shows the results when using the continuous feature alignment (CFA) variant with negative mining.
The hyperparameters (includling the layer at which the representations are extracted) are tuned on the English development set.
The default model is HuBERT Base 
and achieves a 49.9\% P@10 localization performance.
Increasing the scale of the model and data (HuBERT Large trained on 60k hours of English) or changing the architecture (WavLM Base and WavLM Large) degrade performance. %
The only setting in which we observe a slight improvement is for the multilingual variant (mHuBERT), with the precision reaching 50.7\%, but the difference is negligible, despite mHuBERT being trained on 147 languages.
We also evaluate speech representations pretrained on the target language (Hindi) using a wav2vec~2.0 model.
This model achieves a slightly lower localization performance (44.4\% P@10) than HuBERT Base.
However, this difference is likely due to the architecture rather than the pretraining language, as an English wav2vec~2.0 model performs substantially worse (36.7\% P@10).
Overall, while the speech representations matter, the current choice (HuBERT Base) is strong enough.

\subsection{Word-Level Error Analysis}
\label{subsec:errors-word}

Up to %
now we looked at the performance in the aggregate over words in the vocabulary.
We now zoom in on performance per word. 

To understand the impact of the various components,
\Cref{fig:plots} shows how the downstream localization performance (P@10) varies with three factors:
the number of image captions per word,
captioning precision,
and the relative frequency of the most co-occurring word in captions with respect to the query word.
These metrics are computed using the intersection of the three captioners.
For this experiment, we consider the CFA model with negative mining.
We also plot the linear regression fit and 95\% confidence intervals.

First, in \Cref{fig:plots}--left we see that performance correlates %
with the number of captions, indicating that the more frequent a word is the more likely it is to be correctly retrieved.
However, the correlation is weak, especially at the low end:
rare words such as \word{machine}, \word{girl}, or \word{store} perform well even if they have fewer than 50 utterances.
This suggests that large sample counts are not strictly
necessary for good retrieval, offering a promising direction for future work on very low-resource languages.

\Cref{fig:plots}--middle shows a strong positive relationship between performance and captioning precision.
This highlights the importance of precise captions, and partially explains why the cross-lingual setting performs worse that the monolingual one:
cross-lingual precision is worse than monolingual precision (\Cref{tab:alignment-results}).
Notable exceptions are \word{bus}, \word{bed}, \word{bridge},
which have well-aligned captions, but low retrieval scores.
Taking a closer look at these, we observe that the co-occurrences are causing problems for these cases:
\word{bus} (\word{bas}) is always retrieved with the postposition \word{ke} (\eg, \word{bas ke});
\word{bed} (\word{bistar}) is often confused %
with the word \word{room} (\word{kamre}); and %
the word \word{bridge} (\word{pul}, \word{brij}) is confused %
with %
words like \word{above} (\word{upar}) or segmented with auxiliary words like \word{a bridge} (\word{ek pul}).

Finally, \Cref{fig:plots}--right shows that performance correlates inversely with the relative frequency of the most common co-occurring word:
if a word co-occurs often with the query, then performance suffers.
For example, image captioners most often describe lakes or rivers with the phrase \word{body of water} (\eg, \textit{a small shack sits on the edge of a body of water}).
So
\word{body} gets assigned to words such as \word{water} (\word{paani}) or \word{river} (\word{nadi}) rather than its actual translation.

\section{Conclusions}

We showed how spoken captions in a foreign language (Hindi) can be mapped to written words in English using visual grounding.
Our approach extends unsupervised word discovery by incorporating information extracted from the images being described.
Concretely, we split utterances into positive and negative sets based on whether a query keyword appears in the image captions.
The utterances are then aligned, and a spoken segment is retrieved based on how it aligns with each set:
alignment with segments from the positive set serves as positive evidence, while
alignment with segments from the negative set serves as negative evidence.

We find that the alignment based on continuous features achieves the best overall results.
Aligning with discrete features %
performs worse, but is computationally more efficient.
Our best setup outperforms a neural baseline by roughly
39\% absolute for P@10 in keyword localization  and by
44\% for P@10 in keyword spotting.
We show that incorporating negative evidence through negative mining is crucial for good performance and
makes the system robust to the image captioning system.

A gap still remains relative to a topline that uses transcripts for supervision instead of visual grounding.
This is natural, given the inherent variability in how images are described,
but it is further exacerbated by the cultural gap between English and Hindi speakers:
the two groups tend to describe the images differently.
In this work, we used images %
that were of interest to Westerners,
which were then annotated by Hindi speakers.
Future work should %
explore selecting images that are more relevant to the target culture \cite{liu2021visually}, so as to narrow this gap.

Overall, our work demonstrates that visually grounded speech can be used to build cross-lingual vocabularies without textual resources in the target language.
We hope that this approach can help linguists and researchers develop technology that supports and helps preserve all the world's languages~\cite{bird2022acl,bird2024eacl, tapo2024interspeech}.

\bibliographystyle{IEEEtran}
\bibliography{refs}

@string{acl = {Proc. ACL}}

@string{asru = {Proc. ASRU}}

@string{cvpr = {Proc. CVPR}}

@string{eacl = {Proc. EACL}}

@string{eccv = {Proc. ECCV}}

@string{emnlp = {Proc. EMNLP}}

@string{interspeech = {Proc. Interspeech}}

@string{icml = {Proc. ICML}}

@string{icassp = {Proc. ICASSP}}

@string{iclr = {Proc. ICLR}}

@string{lrec = {Proc. LREC}}

@string{jair = {J. Artif. Intell. Res.}}

@string{jmlr = {J. Mach. Learn. Res.}}

@string{jstsp = {IEEE J. Sel. Top. Signal Process.}}

@string{nips = {Proc. NIPS}}

@string{neurips = {Proc. NeurIPS}}

@string{sltu = {Proc. SLTU}}

@string{tmlr = {Trans. Mach. Learn. Res.}}

@string{taslp = {IEEE ACM Trans. Audio Speech Lang. Process.}}

@inproceedings{harwath2015asru,
    author        = {Harwath, David and Glass, James},
    booktitle     = asru,
    title         = {Deep multimodal semantic embeddings for speech and images},
    year          = {2015}
}

@inproceedings{harwath2016nips,
    author        = {Harwath, David and Torralba, Antonio and Glass, James},
    booktitle     = nips,
    title         = {Unsupervised learning of spoken language with visual context},
    year          = {2016}
}

@inproceedings{harwath2018eccv,
    author        = {Harwath, David and Recasens, Adri{\`a} and Sur{\'\i}s, D{\'\i}dac and Chuang, Galen and Torralba, Antonio and Glass, James},
    booktitle     = eccv,
    title         = {Jointly discovering visual objects and spoken words from raw sensory input},
    year          = {2018}
}

@inproceedings{dharwarth2018_interlingua,
    author        = {David Harwath and Galen Chuang and James R. Glass},
    booktitle     = icassp,
    title         = {Vision as an Interlingua: Learning Multilingual Semantic Embeddings of Untranscribed Speech},
    year          = {2018}
}

@inproceedings{kamper2018sltu,
    author        = {Herman Kamper and Michael Roth},
    booktitle     = sltu,
    title         = {Visually Grounded Cross-Lingual Keyword Spotting in Speech},
    year          = {2018}
}

@article{kamper2019taslp,
    author        = {Kamper, Herman and Shakhnarovich, Gregory and Livescu, Karen},
    journal       = taslp,
    title         = {Semantic speech retrieval with a visually grounded model of untranscribed speech},
    volume        = {27},
    year          = {2019}
}

@article{olaleye2022jstsp,
    author        = {Olaleye, Kayode and Oneata, Dan and Kamper, Herman},
    journal       = jstsp,
    title         = {Keyword localisation in untranscribed speech using visually grounded speech models},
    volume        = {16},
    year          = {2022}
}

@article{chrupala2022jair,
  author       = {Grzegorz Chrupa\l{}a},
  title        = {Visually Grounded Models of Spoken Language: {A} Survey of Datasets, Architectures and Evaluation Techniques},
  journal      = jair,
  volume       = {73},
  year         = {2022},
}

@inproceedings{wang2016lrec,
  author       = {Josiah Wang and Robert J. Gaizauskas},
  title        = {Cross-validating Image Description Datasets and Evaluation Metrics},
  booktitle    = lrec,
  year         = {2016},
}

@inproceedings{ohishi2020icassp,
    title        = {Trilingual semantic embeddings of visually grounded speech with self-attention mechanisms},
    author       = {Ohishi, Yasunori and Kimura, Akisato and Kawanishi, Takahito and Kashino, Kunio and Harwath, David and Glass, James},
    booktitle    = icassp,
    year         = {2020},
}

@inproceedings{oneata2024interspeech,
    title        = {Translating speech with just images},
    author       = {Oneata, Dan and Kamper, Herman},
    booktitle    = interspeech,
    year         = {2024},
}

@inproceedings{azuh2019interspeech,
  author       = {Emmanuel Azuh and David Harwath and James R. Glass},
  title        = {Towards Bilingual Lexicon Discovery From Visually Grounded Speech Audio},
  booktitle    = interspeech,
  year         = {2019},
}

@inproceedings{suris2022cvpr,
  author       = {D{\'{\i}}dac Sur{\'{\i}}s and Dave Epstein and Carl Vondrick},
  title        = {Globetrotter: Connecting Languages by Connecting Images},
  booktitle    = cvpr,
  year         = {2022},
}

@inproceedings{sigurdsson2020cvpr,
  author       = {Gunnar A. Sigurdsson and Jean{-}Baptiste Alayrac and Aida Nematzadeh and Lucas Smaira and Mateusz Malinowski and Jo{\~{a}}o Carreira and Phil Blunsom and Andrew Zisserman},
  title        = {Visual Grounding in Video for Unsupervised Word Translation},
  booktitle    = cvpr,
  year         = {2020},
}

@inproceedings{khorrami2021interspeech,
  author       = {Khazar Khorrami and Okko R{\"{a}}s{\"{a}}nen},
  title        = {Evaluation of Audio-Visual Alignments in Visually Grounded Speech Models},
  booktitle    = interspeech,
  year         = {2021},
}

@inproceedings{havard2019icassp,
  author       = {William N. Havard and Jean{-}Pierre Chevrot and Laurent Besacier},
  title        = {Models of Visually Grounded Speech Signal Pay Attention to Nouns: {A} Bilingual Experiment on English and Japanese},
  booktitle    = icassp,
  year         = {2019},
}

@inproceedings{harwath2017acl,
  author       = {David Harwath and James R. Glass},
  title        = {Learning Word-Like Units from Joint Audio-Visual Analysis},
  booktitle    = acl,
  year         = {2017},
}

@article{hubert2021,
author = {Hsu, Wei-Ning and Bolte, Benjamin and Tsai, Yao-Hung Hubert and Lakhotia, Kushal and Salakhutdinov, Ruslan and Mohamed, Abdelrahman},
title = {{HuBERT}: Self-Supervised Speech Representation Learning by Masked Prediction of Hidden Units},
year = {2021},
issue_date = {2021},
publisher = {IEEE Press},
volume = {29},
issn = {2329-9290},
journal = taslp,
numpages = {10}
}

@inproceedings{vanniekerk24_interspeech,
  title     = {Spoken-Term Discovery using Discrete Speech Units},
  author    = {Benjamin {van Niekerk} and Julian Zaïdi and Marc-André Carbonneau and Herman Kamper},
  year      = {2024},
  booktitle = interspeech,
}

@article{interval_piling,
author = {Park, Alex and Glass, James},
year = {2008},
title = {Unsupervised Pattern Discovery in Speech},
volume = {16},
journal = taslp,
doi = {10.1109/TASL.2007.909282}
}

@inproceedings{NIPS2014_19ea3982,
 author = {Zhou, Bolei and Lapedriza, Agata and Xiao, Jianxiong and Torralba, Antonio and Oliva, Aude},
 booktitle = neurips,
 title = {Learning Deep Features for Scene Recognition using Places Database},
 year = {2014}
}

@article{kuchaiev2019nemo,
  title={{NeMo}: A toolkit for building {AI} applications using neural modules},
  author={Kuchaiev, Oleksii and Li, Jason and Nguyen, Huyen and Hrinchuk, Oleksii and Leary, Ryan and Ginsburg, Boris and Kriman, Samuel and Beliaev, Stanislav and Lavrukhin, Vitaly and Cook, Jack and others},
  journal={arXiv preprint arXiv:1909.09577},
  year={2019}
}

@article{scalingspeech2024,
author = {Pratap, Vineel and Tjandra, Andros and Shi, Bowen and Tomasello, Paden and Babu, Arun and Kundu, Sayani and Elkahky, Ali and Ni, Zhaoheng and Vyas, Apoorv and Fazel-Zarandi, Maryam and Baevski, Alexei and Adi, Yossi and Zhang, Xiaohui and Hsu, Wei-Ning and Conneau, Alexis and Auli, Michael},
title = {Scaling speech technology to 1,000+ languages},
year = {2024},
volume = 25,
journal = jmlr,
}

@misc{Honnibal_spaCy_Industrial-strength_Natural_2020,
author = {Honnibal, Matthew and Montani, Ines and Van Landeghem, Sofie and Boyd, Adriane},
doi = {10.5281/zenodo.1212303},
title = {{spaCy}: Industrial-strength Natural Language Processing in {P}ython},
url = {https://doi.org/10.5281/zenodo.1212303},
year = {2020}
}

@inproceedings{huang2023tag2text,
  title={Tag2{T}ext: Guiding Vision-Language Model via Image Tagging},
  author={Huang, Xinyu and Zhang, Youcai and Ma, Jinyu and Tian, Weiwei and Feng, Rui and Zhang, Yuejie and Li, Yaqian and Guo, Yandong and Zhang, Lei},
  booktitle=iclr,
  year={2024}
}

@inproceedings{10.5555/3618408.3619222,
author = {Li, Junnan and Li, Dongxu and Savarese, Silvio and Hoi, Steven},
title = {{BLIP}-2: bootstrapping language-image pre-training with frozen image encoders and large language models},
year = {2023},
booktitle = icml
}

@inproceedings{wang2022gitgenerativeimagetotexttransformer,
      author        = {Jianfeng Wang and Zhengyuan Yang and Xiaowei Hu and Linjie Li and Kevin Lin and Zhe Gan and Zicheng Liu and Ce Liu and Lijuan Wang},
    booktitle       = tmlr,
    title         = {{GIT:} {A} Generative Image-to-text Transformer for Vision and Language},
    year          = {2022}
}

@inproceedings{Plaquet23,
  author={Alexis Plaquet and Hervé Bredin},
  title={{Powerset multi-class cross entropy loss for neural speaker diarization}},
  year=2023,
  booktitle=interspeech,
}

@inproceedings{Olaleye2021AttentionBasedKL,
author = {Olaleye, Kayode and Kamper, Herman},
year = {2021},
booktitle=interspeech,
title = {Attention-Based Keyword Localisation in Speech Using Visual Grounding},
doi = {10.21437/Interspeech.2021-435}
}

@article{smith-waterman,
title = {Identification of common molecular subsequences},
journal = {Journal of Molecular Biology},
volume = {147},
year = {1981},
author = {T.F. Smith and M.S. Waterman}
}

@inproceedings{dunbar2021zero,
  author    = {Dunbar, Erin and Bernard, Myl{\`e}ne and Hamilakis, Nikolaos and Nguyen, Thi-Anh and Seyssel, Marie de and Roz{\'e}, Philippe and Rivi{\`e}re, Morgane and Kharitonov, Eugene and Dupoux, Emmanuel},
  title     = {The Zero Resource Speech Challenge 2021: Spoken Language Modelling},
  booktitle = interspeech,
  year      = {2021},
}

@article{R2018254,
title = {A robust unsupervised pattern discovery and clustering of speech signals},
journal = {Pattern Recognit. Lett.},
volume = {116},
year = {2018},
author = {Kishore Kumar R and Lokendra Birla and Sreenivasa Rao K},
}

@inproceedings{bird2022acl,
  title={Local languages, third spaces, and other high-resource scenarios},
  author={Bird, Steven},
  booktitle=acl,
  year={2022},
}

@inproceedings{bird2024eacl,
  title={Centering the speech community},
  author={Bird, Steven and Yibarbuk, Dean},
  booktitle=eacl,
  year={2024}
}

@inproceedings{tapo2024interspeech,
  title={Leveraging Speech Data Diversity to Document Indigenous Heritage and Culture},
  author={Tapo, Allahsera and Le Ferrand, {\'E}ric and Liu, Zoey and Homan, Christopher and Prud’hommeaux, Emily},
  booktitle=interspeech,
  year={2024}
}

@inproceedings{pirlogeanu2026connectingspeechwordsimages,
    title={Connecting Speech to Words through Images}, 
    author={Gabriel Pirlogeanu and Dan Oneata and Horia Cucu and Herman Kamper},
    year={2026},
    booktitle={Proc. EUSIPCO},
}

@inproceedings{conformer2021,
  author = {Gulati, Anmol and Qin, James and Chiu, Chung-Cheng and Parmar, Niki and Zhang, Yu and Yu, Jiahui and Han, Wei and Wang, Shibo and Zhang, Zhengdong and Wu, Yonghui and Pang, Ruoming},
  booktitle = interspeech,
  title = {Conformer: Convolution-augmented Transformer for Speech Recognition.},
  year = 2020
}

@inproceedings{diwan21_interspeech,
  title     = {{MUCS} 2021: Multilingual and Code-Switching {ASR} Challenges for Low Resource {I}ndian Languages},
  author    = {Anuj Diwan and Rakesh Vaideeswaran and Sanket Shah and Ankita Singh and Srinivasa Raghavan and Shreya Khare and Vinit Unni and Saurabh Vyas and Akash Rajpuria and Chiranjeevi Yarra and Ashish Mittal and Prasanta Kumar Ghosh and Preethi Jyothi and Kalika Bali and Vivek Seshadri and Sunayana Sitaram and Samarth Bharadwaj and Jai Nanavati and Raoul Nanavati and Karthik Sankaranarayanan},
  year      = {2021},
  booktitle = interspeech
}

@inproceedings{hermjakob-etal-2018-box,
    title = "Out-of-the-box Universal {R}omanization Tool uroman",
    author = "Hermjakob, Ulf  and
      May, Jonathan  and
      Knight, Kevin",
    booktitle = acl,
    year = 2018,
}

@inproceedings{qi-etal-2020-stanza,
    title = "{S}tanza: A Python Natural Language Processing Toolkit for Many Human Languages",
    author = "Qi, Peng  and
      Zhang, Yuhao  and
      Zhang, Yuhui  and
      Bolton, Jason  and
      Manning, Christopher D.",
    booktitle = acl,
    year = "2020"
}

@inproceedings{boito2024mhubert,
author={Boito, Marcely Zanon and Iyer, Vivek and Lagos, Nikolaos and Besacier, Laurent and Calapodescu, Ioan},
title={{mHuBERT}-147: A Compact Multilingual {HuBERT} Model},
year=2024,
booktitle=interspeech,
}

@article{Chen2021WavLMLS,
  title={WavLM: Large-Scale Self-Supervised Pre-Training for Full Stack Speech Processing},
  author={Sanyuan Chen and Chengyi Wang and Zhengyang Chen and Yu Wu and Shujie Liu and Zhuo Chen and Jinyu Li and Naoyuki Kanda and Takuya Yoshioka and Xiong Xiao and Jian Wu and Long Zhou and Shuo Ren and Yanmin Qian and Yao Qian and Micheal Zeng and Furu Wei},
  journal=jstsp,
  year={2021},
  volume={16},
}

@inproceedings{wav2vec2,
author = {Baevski, Alexei and Zhou, Henry and Mohamed, Abdelrahman and Auli, Michael},
title = {wav2vec 2.0: a framework for self-supervised learning of speech representations},
year = {2020},
booktitle = nips,
}

@inproceedings{hershcovich2022challenges,
  title={Challenges and strategies in cross-cultural NLP},
  author={Hershcovich, Daniel and Frank, Stella and Lent, Heather and De Lhoneux, Miryam and Abdou, Mostafa and Brandl, Stephanie and Bugliarello, Emanuele and Piqueras, Laura Cabello and Chalkidis, Ilias and Cui, Ruixiang and others},
  booktitle=acl,
  year={2022}
}

@inproceedings{liu2021visually,
  title={Visually grounded reasoning across languages and cultures},
  author={Liu, Fangyu and Bugliarello, Emanuele and Ponti, Edoardo Maria and Reddy, Siva and Collier, Nigel and Elliott, Desmond},
  booktitle=emnlp,
  year={2021}
}

@article{pulikodan2026vaanibenchmarkv10inclusive,
  title={Vaani Benchmark V1. 0: An Inclusive Multimodal Benchmark Dataset for {H}indi},
  author={Pulikodan, Sujith and Basu, Agneedh and Kumar, Saurabh and Bhat, Pranav and Sanka, Visruth and Desai, Nihar and Ghosh, Prasanta Kumar and others},
  journal={arXiv preprint arXiv:2606.21408},
  year={2026}
}

@article{vaani,
  title={{VAANI}: Capturing the language landscape for an inclusive digital {I}ndia},
  author={Pulikodan, Sujith and Singh, Abhayjeet and Basu, Agneedh and Desai, Nihar and Bhat, Pranav D and Dharmaraju, Raghu and Gupta, Ritika and Udupa, Sathvik and Kumar, Saurabh and Sharma, Sumit and others},
  journal={arXiv preprint arXiv:2603.28714},
  year={2026}
}
\end{document}